\documentclass{article}

\usepackage[numbers]{natbib}
\usepackage[dblblindworkshop, final]{neurips_2025}

\usepackage[utf8]{inputenc} 
\usepackage{fontenc}    
\usepackage{hyperref}       
\usepackage{url}            
\usepackage{booktabs}       
\usepackage{amsfonts}       
\usepackage{nicefrac}       
\usepackage{microtype}      
\usepackage{xcolor}         
\usepackage{natbib}

\usepackage{amsmath}
\usepackage{amssymb}
\usepackage{graphicx}
\usepackage{algorithm}
\usepackage{algorithmic}

\hypersetup{
    colorlinks=true,
    linkcolor=blue,
    filecolor=magenta,      
    urlcolor=cyan,
    citecolor=blue,
}

\workshoptitle{NeurIPS 2025 Workshop on Efficient Reasoning}

\title{Causal Reflection with Language Models}

\author{%
  Abi Aryan \\
  Abide AI\\
  \And
  Zac Yung-Chun Liu \\
  Abide AI\\
}

\begin{document}

\maketitle

\begin{abstract}
Large Language Models (LLMs) and traditional Reinforcement Learning (RL) agents lack robust causal reasoning, often relying on spurious correlations. We introduce Causal Reflection, a framework that moves beyond simple reward optimization to build dynamic causal models of an environment. Our approach features a temporal, action-based causal function that models state, action, time, and perturbation to capture delayed and nonlinear effects. We also define a formal \texttt{Reflect} mechanism that identifies mismatches between predicted and observed outcomes, generating causal hypotheses to revise the agent's internal model. Within this architecture, LLMs are not black-box reasoners but structured interpreters, translating formal causal outputs into natural language explanations. This work lays the theoretical groundwork for agents that can adapt, self-correct, and communicate causal understanding.
\end{abstract}

\section{Introduction}

The advancement of artificial intelligence requires systems that understand not just what happens, but why. Traditional reinforcement learning (RL) agents maximize rewards without modeling cause-effect relationships, limiting their ability to adapt or generalize \citep{kiciman2023, seitzer2021}. Similarly, large language models (LLMs), despite their fluency, lack an inherent grasp of causality in dynamic contexts, often failing in novel scenarios \citep{jiao2024, du2017}. While causal reinforcement learning (CRL) offers a promising direction by incorporating causal models, existing methods often assume fixed causal structures, failing to capture the evolving nature of real-world systems where relationships can change over time \citep{he2025, peters2017}. This approach directly confronts critical limitations, preventing agents from succumbing to spurious correlations and enabling them to generalize beyond their training environments \citep{wan2024}.

This paper introduces Causal Reflection, a novel framework that addresses these limitations through three key contributions. First, we define a temporal, action-based causal function capable of modeling both linear and nonlinear causal relationships in dynamic systems. Second, we formalize a \texttt{Reflect} mechanism that enables agents to self-correct their internal causal models by generating and testing hypotheses about prediction errors. Third, we propose a principled integration schema where LLMs act as structured interpreters, translating the formal outputs of our causal model into natural language explanations and counterfactuals.
Our approach represents a shift toward more interpretable, robust, and generalizable AI.

\section{Related Work}

The convergence of causal inference and reinforcement learning is an emerging paradigm that addresses fundamental limitations in traditional sequential decision-making systems. For example, Causal Reinforcement Learning (CRL) seeks to improve policy learning and generalization by embedding causal models, such as Structural Causal Models (SCMs), into the learning process \citep{deng2023, bareinboim2021}.
However, these approaches are often constrained to environments with static, time-invariant causal graphs, limiting their applicability in non-stationary settings \citep{mendez-molina2022, he2025}. While temporal causal models (TSCMs) have been developed to capture time-varying dynamics \citep{gkorgkolis2025, calderon2024}, they typically lack an internal mechanism for an agent to question and revise its own causal assumptions when confronted with structural breaks. These challenges are complicated by foundational issues, as even established methods like Granger causality can fail in complex nonlinear systems \citep{palus2018}, motivating our framework's explicit modeling of perturbations. Concurrently, while LLMs have demonstrated promise on causal benchmarks \citep{kiciman2023}, their reasoning is often shallow and brittle, relying on memorized patterns rather than genuine understanding \citep{wang2024, ashwani2024, chi2024}. This has motivated hybrid approaches that ground LLMs in formal causal structures to enable more robust inference, informing our framework's design \citep{liu2025, gkountouras2024}.

\section{The Causal Reflection Framework}

We formalize a framework for modeling causality in dynamic environments.
Our approach is built on a novel causal function and a self-correction mechanism.

\subsection{The Temporal Action-Based Causal Function}
We define causality over four core components: State ($S_t$), a vector representing the environment's configuration; Action ($A_t$), an intervention on the system; Time ($T_t$), which imposes temporal ordering; and a Perturbation Factor ($\delta$), representing small, unobserved influences that can trigger nonlinear or chaotic effects. The explicit modeling of $\delta$ allows the framework to account for non-stationarity and structural breaks. We formulate a causal function, $C$, that maps these components to a future state. The function progressively incorporates complexity, starting with direct effects, then temporal delays ($k$), and finally the perturbation factor $\delta$. The complete temporal action-based causal function is:
\begin{equation}
C(S_t, A_t, T_t, \delta) \rightarrow S_{t+k}
\end{equation}
This function models systems where causality is not only delayed but also nonlinear and time-varying. The inclusion of $\delta$ allows the function to model how small causes can propagate to produce disproportionate effects, which can be operationalized through forms such as:
\begin{equation}
S_{t+k} = S_t + f(A_t, T_t) \cdot e^{-\delta}
\end{equation}
Here, $f(A_t, T_t)$ is the standard effect of an action, while $e^{-\delta}$ acts as a nonlinear scaling factor, modeling how unforeseen events can amplify or dampen causal effects.

\begin{figure}[h]
    \centering
    \includegraphics[width=0.9\textwidth]{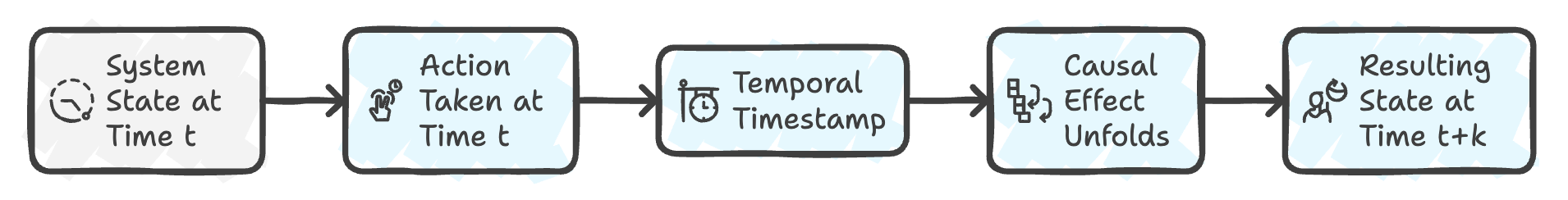} 
    \caption{General workflow of the Causal Reflection framework.
An agent's causal model $C$ makes a prediction, which is compared against the observed outcome. Discrepancies trigger the \texttt{Reflect} mechanism to generate hypotheses and update the model. The LLM translates the formal outputs into natural language.}
    \label{fig:workflow}
\end{figure}

\subsection{The Reflect Mechanism and LLM Integration}
We operationalize self-reflection through a formal \texttt{Reflect} function that enables an agent to learn from discrepancies between predicted and observed outcomes.
When the prediction error $\epsilon = \mathrm{Loss}(\hat{S}_{t+k}, S_{t+k}^{\mathrm{obs}})$ exceeds a threshold $\tau$, a causal mismatch is detected.
The \texttt{Reflect} mechanism is invoked to generate a set of causal hypotheses $H_t$ that could explain the error:
\begin{equation}
    H_t = \mathrm{Reflect}(S_t, A_t, T_t, \delta, \epsilon) := \arg\max_H \left[ P\left(S_{t+k}^{\mathrm{obs}} \mid H\right) - P\left(S_{t+k}^{\mathrm{pred}} \mid C\right) \right]
\end{equation}
Here, $H$ represents a candidate hypothesis, such as a misestimated perturbation factor $\delta$ or an unmodeled confounder. This elevates reflection from a heuristic process to structured causal inference. A detailed formalization of this process is provided in Algorithm \ref{alg:reflect}.

\begin{algorithm}[h!]
\caption{Reflect Mechanism for Causal Hypothesis Generation}
\label{alg:reflect}
\begin{algorithmic}
\REQUIRE Current state $S_t$, action $A_t$, time $T_t$, perturbation $\delta$, causal function $C$, observed outcome $S^{obs}_{t+k}$, loss function $L$, threshold $\tau$
\ENSURE Set of causal hypotheses $H_t$

\STATE $\hat{S}_{t+k} \leftarrow C(S_t, A_t, T_t, \delta)$
\STATE $\epsilon \leftarrow L(\hat{S}_{t+k}, S^{obs}_{t+k})$
\IF{$\epsilon > \tau$}
    \STATE Generate candidate hypotheses $\mathcal{H}$ to explain discrepancy
    \STATE $H_t \leftarrow \arg\max_{H \in \mathcal{H}} \left( \right)$
    \FOR{each $H \in H_t$}
        \STATE Test hypothesis $H$
        \IF{$H$ is valid}
            \STATE Update causal function $C$ with $H$
        \ENDIF
    \ENDFOR
\ELSE
    \STATE $H_t \leftarrow \emptyset$
\ENDIF
\RETURN $H_t$
\end{algorithmic}
\end{algorithm}

This approach represents a significant advance over existing self-reflection paradigms, such as the Reflexion framework \citep{shinn2023}. While Reflexion improves agent behavior through verbal reinforcement and meta-cognitive feedback, its reflection process remains heuristic and unstructured. In contrast, Causal Reflection elevates reflection from intuition to structured causal inference, prompting the agent to generate specific, falsifiable hypotheses about its world model.

Our framework utilizes LLMs not as primary reasoners, but as generative inference engines that translate the formal, symbolic outputs of the causal model into intelligible explanations. The LLM receives the causal tuple $(S_t, A_t, T_t, \delta)$ and the model's prediction, and generates natural language outputs like causal explanations or counterfactuals (e.g., "Had perturbation $\delta$ not occurred, the model predicts the system would have transitioned to state $S'_{t+k}$ instead"). This grounds the LLM's output in a verifiable causal structure, mitigating the risk of ungrounded, "hallucinated" reasoning. A detailed comparison of our framework with traditional RL and CRL is available in Table \ref{tab:comparison}.

\begin{table}[h!]
\centering
\caption{Comparison of Decision-Making Paradigms}
\label{tab:comparison}
\begin{tabular}{@{}llll@{}}
\toprule
\textbf{Dimension} & \textbf{Traditional RL (PPO)} & \textbf{Causal RL (CRL)} & \textbf{Causal Reflection} \\ \midrule
\textbf{Primary Goal} & Maximize cumulative & Improve policy learning & Build an accurate, dynamic \\
& reward.
& (sample efficiency, & causal model for \\
& & generalization) using a & explanation and prediction. \\
& & causal model.
& \\ \addlinespace
\textbf{Core Mechanism} & Policy optimization via & Causal discovery/ & Predictive modeling and \\
& trial-and-error.
& inference on a static & causal hypothesis testing \\
& & world model to inform & on a dynamic world model.
\\
& & policy. & \\ \addlinespace
\textbf{Handling of Time} & Sequential states, but no & Often assumes a static, & Explicitly models temporal \\
& explicit model of & time-invariant causal & delays and time-varying \\
& temporal causality.
& graph. & dynamics. \\ \addlinespace
\textbf{Role of LLM} & N/A, or used for auxiliary & Can be a source of prior & A generative inference \\
& tasks like reward & knowledge for the static & engine that translates the \\
& shaping.
& causal graph. & formal causal model's \\
& & & output into natural language \\
& & & explanations.
\\ \bottomrule
\end{tabular}
\end{table}

\section{Limitations}
Several challenges must be addressed to fully realize the potential of this framework, two of the most urgent and critical ones being-

\textbf{Scalability:} Modeling complex, high-dimensional systems is computationally intensive. The state representation $S_t$ can become prohibitively large, and inferring the causal function C in such spaces is a significant challenge, echoing broader issues in high-dimensional causal inference. Future work should explore factorization and representation learning techniques to create lower-dimensional, causally sufficient state spaces. While reasoning over dynamic causal models increases complexity, the framework is modular: $\delta$ estimation and \textit{Reflect} are invoked conditionally (only when prediction error is high), and causal model updates are localized. For high-dimensional state spaces, we recommend applying dimensionality reduction techniques (e.g., Variational Autoencoders (VAEs) or causal autoencoders) to obtain compact representations.

\textbf{LLM Fidelity and Controllability:} The framework relies on the LLM to be a faithful interpreter of the formal model's output. However, LLMs can "hallucinate" or misrepresent information. Research is needed to develop methods for quantifying and mitigating these "translation errors" to ensure the natural language explanations remain rigorously grounded in the underlying causal inference.
As more experiments and adaptions of Causal Reflection roll in, we are likely to see more limitations discussed.

\section{Discussion and Conclusion}

The Causal Reflection framework offers a path toward more robust, explainable, and aligned AI systems. It has direct applications in domains requiring deep causal understanding, such as human-AI alignment. For example, an agent could hypothesize that a user's "productivity drop" (State) is caused by "unplanned meetings" (Action) with a 24-hour delay, amplified by a "lack of sleep" (Perturbation $\delta$), moving beyond surface-level advice to provide actionable causal insights. 

While this paper establishes the theoretical foundation, a critical next step is developing methods to learn the causal function $C$ from observational and interventional data. We also propose extending the framework to multi-agent systems, where agents must model the causal influence of others' actions \citep{briglia2025}, a nascent but critical research area. We propose validating the framework in simulated environments with known, dynamic causal laws, where an agent's success is measured not by cumulative reward, but by its ability to accurately identify the active causal graph and predict future states, especially after structural breaks.

In conclusion, we introduced Causal Reflection, a framework that shifts the focus of AI agents from reward maximization to building accurate, interpretable causal models of dynamic environments. By formally modeling state, action, time, and perturbations, and by using LLMs as structured interpreters, our approach lays the groundwork for AI systems that can reason, adapt, and explain their understanding of the world.

\bibliographystyle{unsrtnat}
\bibliography{references}

\end{document}